\pdfoutput=1
\documentclass[11pt]{article}

\usepackage[final]{acl}

\usepackage{times}
\usepackage{latexsym}

\usepackage{booktabs}
\usepackage{multirow}
\usepackage{tabularx}
\usepackage{makecell}
\usepackage{amssymb}
\usepackage{pifont}
\usepackage{amsmath}
\usepackage{tcolorbox}
\usepackage{listings}
\usepackage{xcolor}
\tcbuselibrary{skins}
\usepackage{hyperref}

\usepackage[T1]{fontenc}
\usepackage[utf8]{inputenc}

\usepackage{microtype}

\usepackage{inconsolata}

\usepackage{graphicx}

\title{ViDoRAG: Visual Document Retrieval-Augmented Generation\\via Dynamic Iterative Reasoning Agents}

\author{Qiuchen Wang$^{1}\thanks{This work was done during an internship at Tongyi Lab, Alibaba Group. \quad qiuchenwang@mail.ustc.edu.cn}$,
Ruixue Ding$^{2}$,
Zehui Chen$^{1}$,
Weiqi Wu$^{3}$,
Shihang Wang$^{2}$,\\ 
\textbf{Pengjun Xie$^{2}$, 
Feng Zhao$^{1}\thanks{Corresponding author}$}\\
$^{1}$MoE Key Laboratory of Brain-inspired Intelligent Perception and Cognition, USTC\\
$^{2}$Tongyi Lab, Alibaba Group \quad
$^{3}$Shanghai Jiao Tong University\\ 
\texttt{Dataset \& Code: \url{https://github.com/Alibaba-NLP/ViDoRAG}}
}

\author{Qiuchen Wang,
Ruixue Ding,
Zehui Chen,
Weiqi Wu,\\
\textbf{Shihang Wang,
Pengjun Xie, 
Feng Zhao$\thanks{Corresponding author}$}\\
\includegraphics[height=0.4cm]{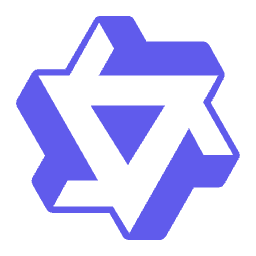} Tongyi Lab, Alibaba Group \quad\\
\texttt{Dataset \& Code: \url{https://github.com/Alibaba-NLP/ViDoRAG}}
}

\begin{document}
\maketitle

\begin{abstract}
Understanding information from visually rich documents remains a significant challenge for traditional Retrieval-Augmented Generation (RAG) methods. Existing benchmarks predominantly focus on image-based question answering (QA), overlooking the fundamental challenges of efficient retrieval, comprehension, and reasoning within dense visual documents. To bridge this gap, we introduce \textbf{ViDoSeek}, a novel dataset designed to evaluate RAG performance on visually rich documents requiring complex reasoning. Based on it, we identify key limitations in current RAG approaches: (i) purely visual retrieval methods struggle to effectively integrate both textual and visual features, and (ii) previous approaches often allocate insufficient reasoning tokens, limiting their effectiveness. To address these challenges, we propose \textbf{ViDoRAG}, a novel multi-agent RAG framework tailored for complex reasoning across visual documents. ViDoRAG employs a Gaussian Mixture Model (GMM)-based hybrid strategy to effectively handle multi-modal retrieval. To further elicit the model's reasoning capabilities, we introduce an iterative agent workflow incorporating exploration, summarization, and reflection, providing a framework for investigating test-time scaling in RAG domains. Extensive experiments on ViDoSeek validate the effectiveness and generalization of our approach. Notably, ViDoRAG outperforms existing methods by over 10\% on the competitive ViDoSeek benchmark. 
\end{abstract}
\section{Introduction}

\begin{figure}[!t]
    \centering 
    \includegraphics[width=1.01\columnwidth]{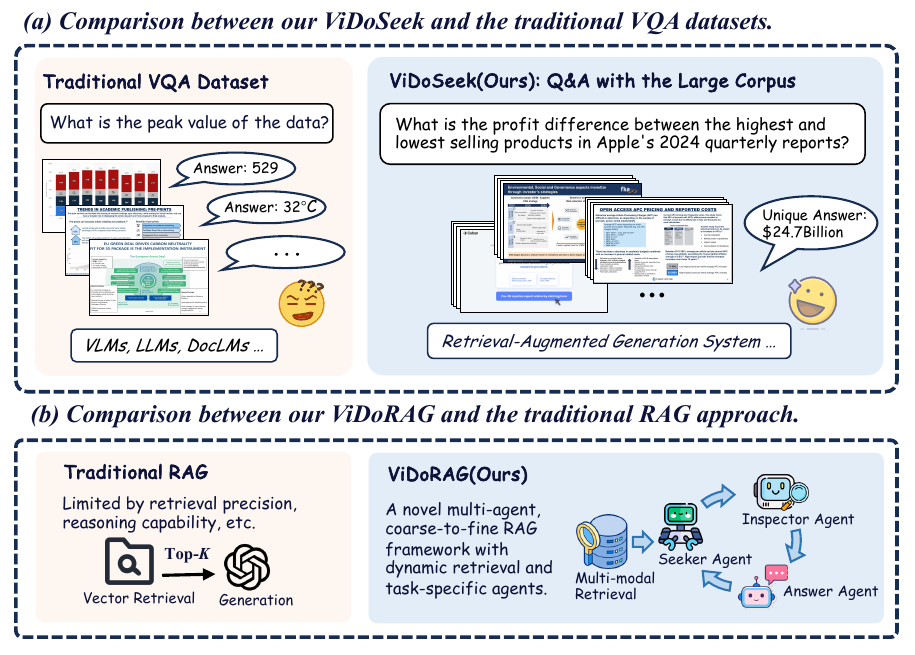}
    \caption{
    Comparison of our work with the existing datasets and methods.
    (a) In traditional datasets, each query must be paired with specific images or documents. In our ViDoSeek, each query can obtain a unique answer within the large corpus.
    (b) Our ViDoRAG is a multi-agent, coarse-to-fine framework specifically optimized for visually rich documents.
    }
    \label{fig:compare_data_rag}
\end{figure}

Retrieval-Augmented Generation (RAG) enhances Large Models (LMs) by enabling them to use external knowledge to solve problems. As the expression of information becomes increasingly diverse, we often work with visually rich documents that contain diagrams, charts, tables, etc. These visual elements make information easier to understand and are widely used in education, finance, law, and other fields. Therefore, researching RAG within visually rich documents is highly valuable.

In practical applications, RAG systems often need to retrieve information from a large collection consisting of hundreds of documents, amounting to thousands of pages. As shown in Fig. \ref{fig:compare_data_rag}, existing Visual Question Answering (VQA) benchmarks aren't designed for such large corpus. The queries in these benchmarks are typically paired with one single image\cite{methani2020plotqa,masry2022chartqa,li2024multimodal,mathew2022infographicvqa} or document\cite{ma2024mmlongbenchdocbenchmarkinglongcontextdocument}, which is used for evaluating Q\&A tasks but not suitable for evaluating RAG systems. The answers to queries in these datasets may not be unique within the whole corpus.

To address this gap, we introduce ViDoSeek, a novel dataset designed for visually rich document retrieval-reason-answer. In ViDoSeek, each query has a unique answer and specific reference pages. It covers the diverse content types and multi-hop reasoning that most VQA datasets include.  This specificity allows us to better evaluate retrieval and generation performance separately. 

Moreover, to enable models to effectively reason over a large corpus, we propose ViDoRAG, a multi-agent, coarse-to-fine retrieval-augmented generation framework tailored for visually rich documents. Our approach is based on two critical observations:
\textbf{(i) Inefficient and Variable Retrieval Performance.} 
Traditional OCR-based retrieval struggles to capture visual information. With the development of vision-based retrieval, it is easy to capture visual information\cite{faysse2024colpali,yu2024visrag,zhai2023sigmoid}. However, there lack of an effective method to integrate visual and textual features, resulting in poor retrieval of relevant content.
\textbf{(ii) Insufficient Activation of Reasoning Capabilities during Generation.} 
Previous studies on inference scaling for RAG focus on expanding the length of retrieved documents\cite{jiang2024longrag,shao2025scaling,xu2023retrieval}. However, due to the characteristics of VLMs, only emphasizing on the quantity of knowledge without providing further reasoning guidance presents certain limitations. There is a need for an effective inference scale-up method to efficiently utilize specific action spaces, such as resizing and filtering, to fully activate reasoning capabilities.

Building upon these insights, ViDoRAG introduces improvements in both retrieval and generation. We propose Multi-Modal Hybrid Retrieval, which combines both visual and textual features and dynamically adjusts results distribution based on Gaussian Mixture Models (GMM) prior. This approach achieves the optimal retrieval distribution for each query, enhancing generation efficiency by reducing unnecessary computations.
During generation, our framework comprises three agents: the seeker, inspector, and answer agents. The seeker rapidly scans thumbnails and selects relevant images with feedback from the inspector. The inspector reviews, then provides reflection and offers preliminary answers. The answer agent ensures consistency and gives the final answer. This framework reduces exposure to irrelevant information and ensures consistent answers across multiple scales.

Our major contributions are as follows:
\begin{itemize}
    \item We introduce ViDoSeek, a benchmark specifically designed for visually rich document retrieval-reason-answer, fully suited for evaluation of RAG within large document corpus.
    \item We propose ViDoRAG, a novel RAG framework that utilizes a multi-agent, actor-critic paradigm for iterative reasoning, enhancing the noise robustness of generation models. 
    \item We introduce a GMM-based multi-modal hybrid retrieval strategy to effectively integrate visual and textual pipelines.
    \item Extensive experiments demonstrate the effectiveness of our method. ViDoRAG significantly outperforms strong baselines, achieving over 10\% improvement, thus establishing a new state-of-the-art on ViDoSeek.
\end{itemize}
\section{Related Work}

\paragraph{Visual Document Q\&A Benchmarks.}
Visual Document Question Answering is focused on answering questions based on the visual content of documents\cite{antol2015vqa,ye2024mplug,wang2024leave}. While most existing research \cite{methani2020plotqa, masry2022chartqa, li2024multimodal, mathew2022infographicvqa} has primarily concentrated on question answering from single images, recent advancements have begun to explore multi-page document question answering, driven by the increasing context length of modern models \cite{mathew2021docvqa, ma2024mmlongbenchdocbenchmarkinglongcontextdocument, tanaka2023slidevqa}. However, prior datasets were not well-suited for RAG tasks involving large collections of documents. To fill this gap, we introduce ViDoSeek, the first large-scale document collection QA dataset, where each query corresponds to a unique answer across a collection of $\sim 6k$ images.

\begin{figure*}[t]
    \centering 
    \includegraphics[width=1.\textwidth]{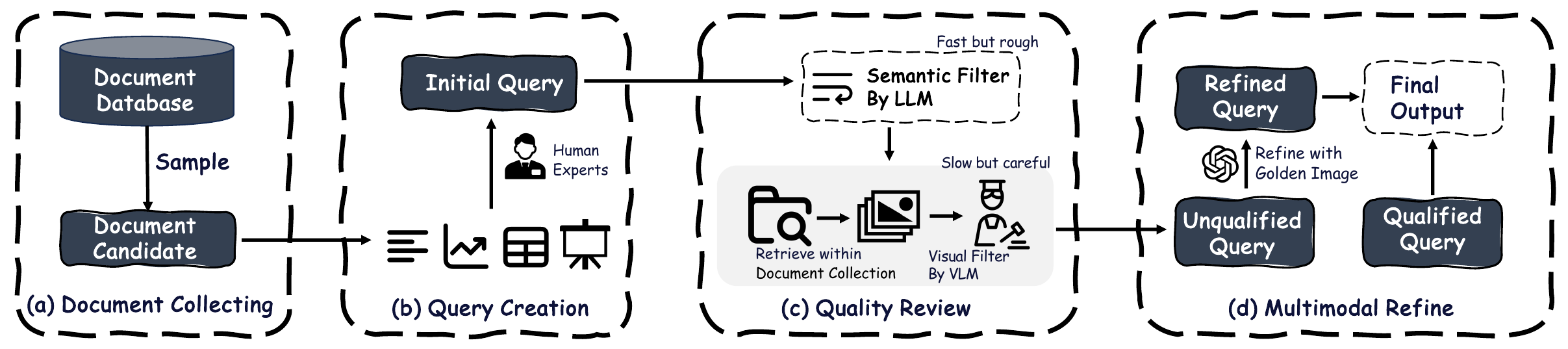}
    \caption{\textbf{Data Construction pipeline.} (a) We sample and filter documents according to the requirements to obtain candidates. (b) Then experts construct the initial query from different contents. (c) After that, we prompt GPT-4 to directly determine whether the query is a general query. The remaining queries are carefully reviewed with top-\textit{K} recall images. (d) Finally, unqualified queries are refined paired with golden image by GPT-4o.}
    \label{fig:construction_pipeline}
\end{figure*}

\paragraph{Retrieval-augmented Generation.}
With the advancement of large models, RAG has enhanced the ability of models to incorporate external knowledge \cite{lewis2020retrieval,chen2024mindsearch,wu2025webwalker}. In prior research, retrieval often followed the process of extracting text via OCR technology \cite{chen2024bge,lee2024nv,robertson2009probabilistic}. Recently, the growing interest in multimodal embeddings has greatly improved image retrieval tasks \cite{faysse2024colpali,yu2024visrag}. 
Additionally, there are works that focus on In-Context Learning in RAG\cite{agarwal2025many,yue2024inference,team2024gemini,weijia2023replug}.
Our work builds upon these developments by combining multi-modal hybrid retrieval with a coarse-to-fine multi-agent generation framework, seamlessly integrating various embedding and generation models into a scalable framework.

\section{Problem Formulation}
\label{sec:problem_define}
Given a query as $q$, and we have a collection of documents $\mathcal{C} = \{ \mathcal{D}_1, \mathcal{D}_2, \ldots, \mathcal{D}_M \}$ which contains $M$ documents. Each document $\mathcal{D}_m$ consists of $N$ pages, each image representing an individual page, defined as $\mathcal{D}_m = \{ \mathbf{I}_1, \mathbf{I}_2, \ldots, \mathbf{I}_N \}$. The total number of images included in the collection is $\sum_{m=1}^{M} |\mathcal{D}_m|$. We aim to retrieve the most relevant information efficiently and accurately and generate the final answer $a$ to the query $q$.

\section{ViDoSeek Dataset}
\label{sec:dataset}

Existing VQA datasets typically consist of queries paired with a single image or a few images. However, in practical application scenarios, users often pose questions based on a large-scale corpus rather than targeting an individual document or image. To better evaluate RAG systems, we prefer questions that have unique answers when retrieving from a large corpus. 
To address this need, we introduce a novel \textbf{Vi}sually rich \textbf{Do}cument dataset specifically designed for RAG systems, called ViDoSeek. 
Below we provide the pipeline for constructing the dataset(\S \ref{sec:data_pipeline})  and a detailed analysis of the dataset(\S \ref{sec:data_analysis}).

\subsection{Dataset Construction.}
\label{sec:data_pipeline}
To construct the ViDoSeek dataset, we developed a four-step pipeline to ensure that the queries meet our stringent requirements. As illustrated in Figure \ref{fig:construction_pipeline}, our dataset comprises two parts: one annotated from scratch by our AI researchers, and the other derived from refining queries in the existing open-source dataset SlideVQA \cite{tanaka2023slidevqa}. 
For the open-source dataset, we initiate the query refinement starting from the third step of our pipeline. For the dataset we build from scratch, we follow the entire pipeline beginning with document collection. The following outlines our four-step pipeline:
\paragraph{Step 1. Document Collecting.}
As slides are a widely used medium for information transmission today, we selected them as our document source. We began by collecting English-language slides containing 25 to 50 pages, covering 12 domains such as economics, technology, literature, and geography. And we filtered out 300 slides that simultaneously include text, charts, tables, and two-dimensional layouts which refer to flowcharts, diagrams, or any visual elements composed of various components and are a distinctive feature of slides.
\begin{table*}[!t]
    \small
    \centering
    \caption{\textbf{Comparison of existing dataset with ViDoSeek.}}
    \resizebox{1.0\textwidth}{!}{
    \label{tab:data_compare}
    \begin{tabular}{lcccc}
    \toprule
    \textsc{\textbf{Dataset}} & \textsc{\textbf{Domain}} & \textsc{\textbf{Content Type}} & \textsc{\textbf{Reference Type}} & \textsc{\textbf{Large Document Collection}} \\
    \midrule
    PlotQA\cite{methani2020plotqa} & Academic & Chart & Single-Image & \ding{55} \\
    ChartQA\cite{masry2022chartqa} & Academic & Chart & Single-Image & \ding{55} \\
    ArxivQA\cite{li2024multimodal} & Academic & Chart & Single-Image & \ding{55} \\
    InfoVQA\cite{mathew2022infographicvqa} & Open-Domain & Text, Chart, Layout & Single-Image & \ding{55} \\
    DocVQA\cite{mathew2021docvqa} & Open-Domain & Text, Chart, Table & Single-Document & \ding{55} \\
    MMLongDoc\cite{ma2024mmlongbenchdocbenchmarkinglongcontextdocument} & Open-Domain & Text, Chart, Table, Layout & Single-Document & \ding{55} \\
    SlideVQA\cite{tanaka2023slidevqa} & Open-Domain & Text, Chart, Table, Layout & Single-Document & \ding{55} \\
    \midrule
    \textbf{ViDoSeek(Ours)} & Open-Domain & Text, Chart, Table, Layout & Multi-Documents & \ding{51} \\
    \bottomrule
    \end{tabular}
}

\end{table*}

\paragraph{Step 2. Query Creation.}

To make the queries more suitable for RAG over a large-scale collection, our experts were instructed to construct queries that are specific to the document.
Additionally, we encouraged constructing queries in various forms and with different sources and reasoning types to better reflect real-world scenarios. 

\paragraph{Step 3. Quality Review.}

In large-scale retrieval and generation tasks, relying solely on manual annotation is challenging due to human brain limitations. To address this, we propose a review module that automatically identifies problematic queries.

\paragraph{Step 4. Multimodal Refine.}
% refine问题 同时保证问题中不含答案
In this final step, we refine the queries that did not meet our standards during the quality review. We use carefully designed VLM-based agents to assist us throughout the entire dataset construction pipeline.

\subsection{Dataset Analysis}
\label{sec:data_analysis}

\paragraph{Dataset Statistics.} ViDoSeek is the first dataset specifically designed for question-answering over large-scale document collections. It comprises approximately $\sim 1.2k$ questions across a wide array of domains, addressing four key content types: Text, Chart, Table, and Layout. Among these, the Layout type poses the greatest challenge and represents the largest portion of the dataset. Additionally, the queries are categorized into two reasoning types: single-hop and multi-hop. Further details of the dataset can be found in the Appendix \ref{appendix:dataset_composition} and \ref{appendix:data_construction_pipeline}.

\paragraph{Comparative Analysis.}
Table \ref{tab:data_compare} highlights the limitations of existing datasets, which are predominantly tailored for scenarios involving single images or documents, lacking the capacity to handle the intricacies of retrieving relevant information from large collections. ViDoSeek bridges this gap by offering a dataset that more accurately mirrors real-world scenarios. This facilitates a more robust and scalable evaluation of RAG systems.

\begin{figure*}[!t]
    \centering 
    \includegraphics[width=1\textwidth]{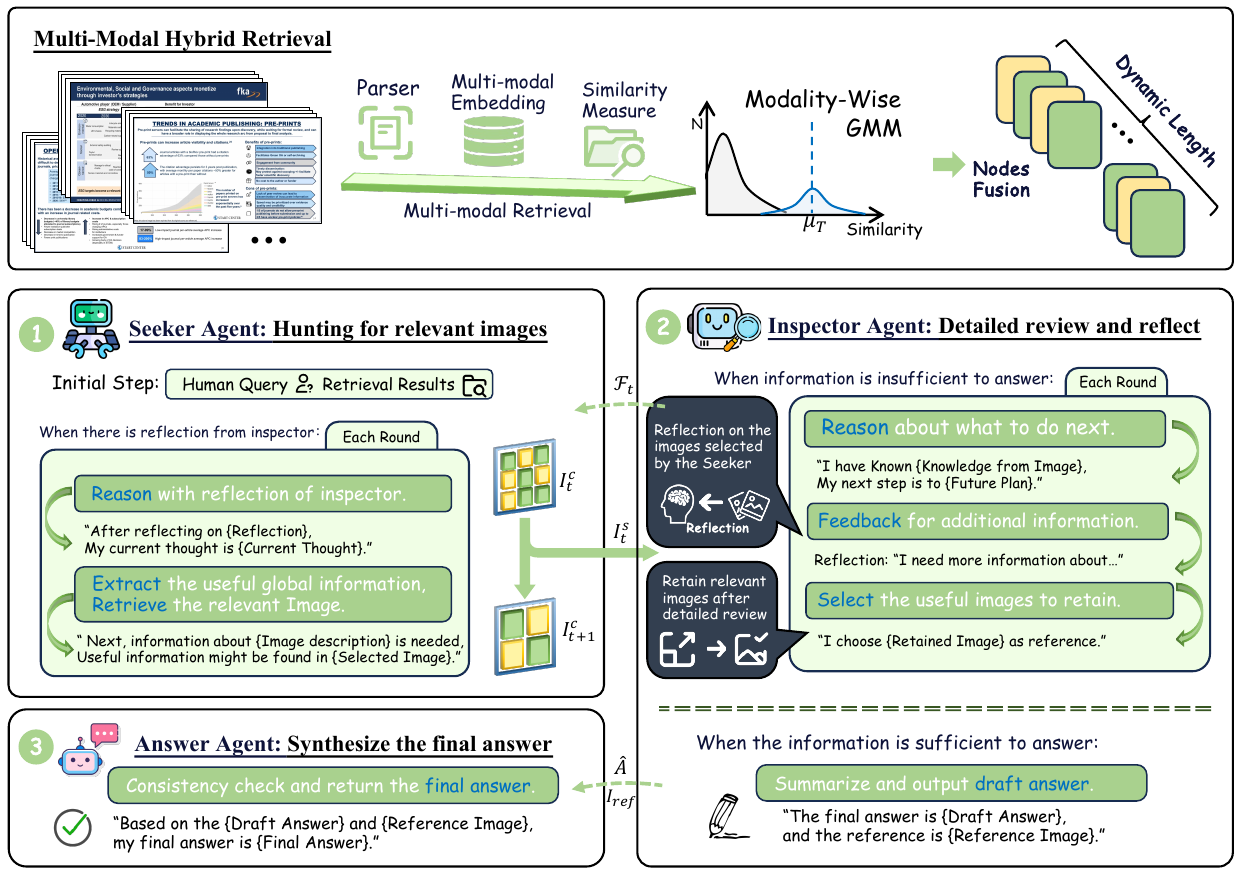}
    \caption{\textbf{ViDoRAG Framework.}}
    \label{fig:pipeline}
\end{figure*}

\section{Method}
In this section, drawing from insights and foundational ideas, we present a comprehensive description of our \textbf{ViDoRAG} framework, which integrates two modules: Multi-Modal Hybrid Retrieval (\S \ref{sec:ret}) and Multi-Scale View Generation (\S \ref{sec:gen}).

\subsection{Multi-Modal Hybrid Retrieval}
\label{sec:ret}
For each query, our approach involves retrieving information through both textual and visual pipelines, dynamically determining the optimal value of top-K using a Gaussian Mixture Model (GMM), and merging the retrieval results from both pipelines.
\paragraph{Adaptive Recall with Gaussian Mixture Model.}
Traditional methods rely on a static hyperparameter, $\mathcal{K}$, to retrieve the top-\textit{K} images or text chunks from a corpus. A smaller $\mathcal{K}$ might fail to capture sufficient references needed for accurate responses, as the most relevant nodes are not always ranked at the top. Conversely, a larger $\mathcal{K}$ can slow down inference and introduce inaccuracies due to noise. Additionally, manually tuning $\mathcal{K}$ for different scenarios is troublesome.

Our objective is to develop a straightforward yet effective method to automatically determine $\mathcal{K}$ for each modality, without the dependency on a fixed value. We utilize the similarity $\mathcal{S}$ of the embedding $E$ to quantify the relevance between the query and the document collection $\mathcal{C}$:
\begin{equation}
\mathcal{S}(q, \mathcal{C}) = \{s_i | cos(E_q,E_{p_i}) , p_i \in \mathcal{C}\}
\end{equation}
where $s_i$ represents the cosine similarity between the query $\mathcal{Q}$ and page $p_i$. In the visual pipeline, a page corresponds to an image, whereas in the textual pipeline, it corresponds to chunks of OCR text. We propose that the distribution of $\mathcal{S}$ follows a GMM and we consider they are sampled from a bimodal distribution $\mathcal{P}(s)$ shown in Fig.\ref{fig:pipeline}:
\begin{equation}
\small \mathcal{P}(s) = w_F \cdot \mathcal{N}(s \mid \mu_F, \sigma_F^2) + w_T \cdot \mathcal{N}(s \mid \mu_T, \sigma_T^2) \label{eq:gaussian}
\end{equation}
where $\mathcal{N}$ represents a Gaussian distribution, with $w,\mu,\sigma^2$ indicating the weight, mean, and variance, respectively. The subscripts $T$ and $F$ refer to the distributions of pages with high and low similarity. The distribution with higher similarity is deemed valuable for generation. The Expectation-Maximization (EM) algorithm is utilized to estimate the prior probability $\mathcal{P}(T|s, \mu_T, \sigma_T^2)$ for each modality. The dynamic value of $\mathcal{K}$ is defined as:
\begin{equation}
\mathcal{K} = | \{ p_i \in \mathcal{C} \mid p_i \sim \mathcal{N}(\mu_T, \sigma_T^2) \} |
\end{equation}

Considering that the similarity score distribution for different queries within a document collection may not strictly follow a standard distribution, we establish upper and lower bounds to manage outliers. The EM algorithm is employed sparingly, less than $\sim 1\%$ of the time. Dynamically adjusting $\mathcal{K}$ enhances generation efficiency compared to a static setting. Detailed analysis is available in \S \ref{sec:analysis:time}.

\paragraph{Textual and Visual Hybrid Retrieval.}
In the previous step, nodes were retrieved from both pipelines. In this phase, we integrate them:
\begin{equation}
\mathcal{R}_{hybrid} = Sort[\mathcal{F}(\mathcal{R}_{Text},\mathcal{R}_{Visual})]
\end{equation}
where $\mathcal{R}_{Text}$ and $\mathcal{R}_{Visual}$ denote the retrieval results from the textual and visual pipelines, respectively. The function $\mathcal{F}(\cdot)$ signifies a union operation, and $Sort(\cdot)$ arranges the nodes in their original sequence, as continuous pages often exhibit correlation \cite{yu2024defense}. 

The textual and visual retrieval pipelines demonstrate varying levels of performance for different features. Without adaptive recall, the combined retrieval $\mathcal{R}_{hybrid}$ can become excessive. Adaptive recall ensures that effective retrievals are concise, while traditional pipelines yield longer recall results. This strategy optimizes performance relative to context length, underscoring the value of adaptive recall in hybrid retrieval.

\subsection{Multi-Agent Generation with Iterative Reasoning}
\label{sec:gen}
During the generation, we introduce a multi-agent framework which consists of three types of agents: the Seeker Agent, the Inspector Agent, and the Answer Agent. As illustrated in Fig. \ref{fig:pipeline}, this framework extracts clues, reflects, and answers in a coarse-to-fine manner from a multi-scale perspective. More details are provided in Appendix \ref{appendix: gen}.

\paragraph{Seeker Agent: Hunting for relevant images.} 
The Seeker Agent is responsible for selecting from a coarse view and extracting global cues based on the query and reflection from the Inspector Agent. We have made some improvements to ReAct\cite{yao2022react} to facilitate better memory management. 
The action space is defined as the selection of the images. Initially, the agent will reason only based on the query $\mathcal{Q}$ and select the most relevant images $\mathbf{I}^{\text{s}}_0$ from the candidate images $\mathbf{I}^{\text{c}}_0$, while the initial memory $\mathcal{M}_0$ is empty.
In step $t$, the candidate images $\mathbf{I}^{\text{c}}_{t + 1}$ are the complement of previously selected images $\mathbf{I}^{\text{s}}_{t}$, defined as $\mathbf{I}^{\text{c}}_{t + 1}=\mathbf{I}^{\text{c}}_{t}\setminus\mathbf{I}^{\text{s}}_{t}$.
The seeker has received the reflection $\mathcal{F}_{t - 1}$ from the inspector, which includes an evaluation of the selected images and a more detailed description of the requirements for the images. The Seeker integrates feedback $\mathcal{F}_{t - 1}$ from the Inspector, which includes an evaluation of the selected images and a description of image requirements, to further refine the selection $\mathbf{I}^{s}_{t}$ and update the memory $\mathcal{M}_{t+1}$:
\begin{equation}
\mathbf{I}^{c}_{t+1},~\mathcal{M}_{t+1} = \Theta(\mathbf{I}^{c}_{t},\mathcal{Q}, \mathcal{M}_{t}, \mathcal{F}_{t-1})
\end{equation}
where $\mathcal{M}_{t+1}$ represents the model's thought content in step $t$ under the ReAct paradigm, maintaining a constant context length. The process continues until the Inspector determines that sufficient information is available to answer the query, or the Seeker concludes that no further relevant images exist among the candidates.

\paragraph{Inspector Agent: Review in detail and Reflect.}
In baseline scenarios, increasing the top-$K$ value improves recall@$K$, but accuracy initially rises and then falls. This is attributed to interference from irrelevant images, referred to as noise, affecting model generation. To address this, we use Inspector to perform a more fine-grained inspection of the images. In each interaction with the Seeker, the Inspector's action space includes providing feedback or drafting a preliminary answer. 
At step $t$, the inspector reviews images at high resolution, denoted as $\Theta(\mathbf{I}^c_t \cup \mathbf{I}^{r}_{t-1}, \mathcal{Q})$ where $\mathbf{I}^{r}_{t-1}$ are images retained from the previous step and $\mathbf{I}^{c}_{t}$ are from the Seeker.
If the current information is sufficient to answer the query, a draft answer $\hat{\mathcal{A}}$ is provided, alongside a reference to the relevant image:
\begin{equation}
\hat{\mathcal{A}},~\mathbf{I}^{ref} = \Theta(\mathbf{I}^c_t \cup \mathbf{I}^{r}_{t-1}, \mathcal{Q})
\end{equation}
Conversely, if more information is needed, the Inspector offers feedback $\mathcal{F}_{t}$ to guide the Seeker in better image selection and identifies images $\mathbf{I}^r_t$ to retain for further review in the next step $t+1$:
\begin{equation}
    \mathcal{F}_t,~\mathbf{I}^r_t = \Theta(\mathbf{I}^c_t \cup \mathbf{I}^{r}_{t-1}, \mathcal{Q})
\end{equation}
The number of images the Inspector reviews is typically fewer than the Seeker's, ensuring robustness in reasoning, particularly for Visual Language Models with moderate reasoning abilities.

\begin{table*}[!ht]
    \small
    \centering
    \caption{\textbf{Overall Generation performance.}}
    \resizebox{0.80\textwidth}{!}{
    \label{tab:overall}
    \begin{tabular}{l|cc|cccc|c}
    \toprule
    \multirow{2}{*}{\textsc{\textbf{Method}}} & 
    \multicolumn{2}{c|}{\textsc{\textbf{Reasoning Type}}} &
    \multicolumn{4}{c|}{\textsc{\textbf{Answer Type}}} &
    \multicolumn{1}{c}{\multirow{2}{*}{\textsc{\textbf{Overall}}}} \\
    & \textbf{Single-hop} & \textbf{Multi-hop} & \textbf{Text} & \textbf{Table} & \textbf{Chart} & \textbf{Layout} & \\
    \midrule
    \multicolumn{8}{c}{$\textit{Llama3.2-Vision-90B-Instruct}$}\\
    \midrule
    Upper Bound & 83.1 & 78.7 & 88.7 & 73.1 & 68.1 & 85.1 & 81.1 \\
    \midrule
    TextRAG & 42.6 & 45.7 & 67.6 & 41.8 & 25.4 & 45.9 & 43.9  \\
    VisualRAG & 61.8 & 60.5 & 82.5 & 48.5 & 52.2 & 63.9 & 61.2 \\
    ViDoRAG (\textbf{Ours}) & 73.3 & 68.5 & 85.1 & 65.6 & 56.1 & 74.7 & 71.2 \\
    \midrule
    \multicolumn{8}{c}{$\textit{Qwen2.5-VL-7B-Instruct}$}\\
    \midrule
    Upper Bound & 77.5 & 78.2 & 88.4 & 77.1 & 69.4 & 78.8 & 77.9 \\
    \midrule
    TextRAG & 59.6 & 55.7 & 78.7 & 53.8 & 40.7 & 60.5 & 57.6 \\
    VisualRAG & 66.8 & 64.3 & 84.9 & 61.1 & 52.8 & 67.5 & 65.7 \\
    ViDoRAG (\textbf{Ours}) & 70.4 & 67.3 & 81.9 & 65.2 & 57.7 & 71.3 & 69.1 \\
    \midrule
    \multicolumn{8}{c}{$\textit{GPT-4o (Closed-Sourced Models)}$}\\
    \midrule
    Upper Bound & 88.8 & 86.3 & 97.5 & 85.7 & 77.1 & 89.4 & 87.7 \\
    \midrule
    TextRAG & 64.3 & 62.6 & 78.7 & 61.0 & 48.4 & 66.1 & 63.5\\
    VisualRAG  & 75.7 & 66.1 & 90.1 & 62.4& 58.5 & 75.4 & 72.1 \\
    ViDoRAG (\textbf{Ours}) & 83.5 & 74.1 & 88.5 & 73.6 & 76.4 & 80.4 & 79.4 \\
    \bottomrule
    \end{tabular}
}
\end{table*}

\paragraph{Answer Agent: Synthesize the final answer.}~ In our framework, the Seeker and Inspector engage in a continuous interaction, and the answer agent provides the answer in the final step. To balance accuracy and efficiency, the Answer Agent verifies the consistency of the Inspector's draft answer $\hat{\mathcal{A}}$. If the reference image matches the Inspector's input, the draft answer is accepted as the final answer $\mathcal{A}=\hat{\mathcal{A}}$.
If the reference image is a subset of the input image, the answer agent should check for consistency between the draft answer $\hat{\mathcal{A}}$ and the reference image, then give the final answer $\mathcal{A}$: 
 If the reference image is a subset of Inspector's the input, the Answer Agent ensures consistency between the draft answer $\hat{\mathcal{A}}$ and the reference image before finalizing the answer $\mathcal{A}$:
\begin{equation}
\mathcal{A} = \Theta(\mathbf{I}_{ref}, \mathcal{Q}, \hat{\mathcal{A}})
\end{equation}
The Answer Agent utilizes the draft answer as prior knowledge to refine the response from coarse to fine. The consistency check between the Answer Agent and Inspector Agent enhances the depth and comprehensiveness of the final answer.
\section{Experiments}

\subsection{Experimental Settings}
\paragraph{Evaluation Metric} For our end-to-end evaluation, we employed a model-based assessment using GPT-4o, which involved assigning scores from 1 to 5 by comparing the reference answer with the final answer. Answers receiving scores of 4 or above were considered correct, and we subsequently calculate accuracy as the evaluation metric. For retrieval evaluation, we use recall as the metric.

\paragraph{Baselines and Oracle.} 
We selecte Nv-embed-V2\cite{lee2024nv} and ColQwen2\cite{faysse2024colpali} as the retrievers for the TextRAG and VisualRAG baselines, respectively. Based on their original settings, we choose the top-5 recall results as the generation input, which equals the average length of dynamic recall results. This ensures a fair comparison and highlights the advantages of our method. The \textbf{Oracle} serves as the upper bound performance, where the model responds based on the golden page without retrieval or other operations. 

\subsection{Main Results}
As shown in Table. \ref{tab:overall}, we conducted experiments on both closed-source and open-source models: GPT-4o, Qwen2.5-7B-Instruct, Qwen2.5-VL-7B\cite{yang2024qwen2}-Instruct, Llama3.2-Vision-90B-Instruct. Closed-source models generally outperform open-source models performance. 
It is worth mentioning that the qwen2.5-VL-7B has shown excellent instruction-following and reasoning capabilities within our framework. In contrast, we found that the llama3.2-VL requires 90B parameters to accomplish the same instructions, which may be related to the model's pre-training domain.
The results suggest that while API-based models offer strong baseline performance, our method is also effective in enhancing the performance of open-source models, offering promising potential for future applications. 
To further demonstrate the robustness of the framework, we constructed a pipeline using data to rewrite queries from SlideVQA\cite{tanaka2023slidevqa}, making the queries suitable for scenarios involving large corpora. The experimental results are presented the analysis.

\begin{table}[!t]
    \small
    \centering
    \caption{Retrieval Performance on ViDoSeek.}
    \resizebox{1\linewidth}{!}{
    \label{tab:ret}
    \begin{tabular}{lcccc}
    \toprule
    \textbf{Retriever} & \textbf{Recall@1} & \textbf{Recall@3} & \textbf{Recall@5} & \textbf{MRR@5}\\
    \midrule
    BM25 & 55.2 & 77.4 & 84.5 & 66.5 \\
    BGE-M3\cite{chen2024bge} & 60.2 & 79.3 & 87.6 & 70.5\\
    NV-Embed-V2\cite{lee2024nv} & 64.1 & 83.5  & 90.3 & 74.7 \\
    \midrule
    VisRAG-Ret\cite{yu2024visrag} & 64.4 & 84.1 & 91.2 & 75.2\\
    ColPali\cite{faysse2024colpali} & 70.6 & 87.9 & 92.8 & 79.6\\
    ColQwen2\cite{faysse2024colpali} & 75.4 & 89.7 & 95.1 & 83.3 \\
    \bottomrule
    \end{tabular}
}
\end{table}
\begin{figure}[!h]
    \centering 
    \includegraphics[width=0.95\linewidth]{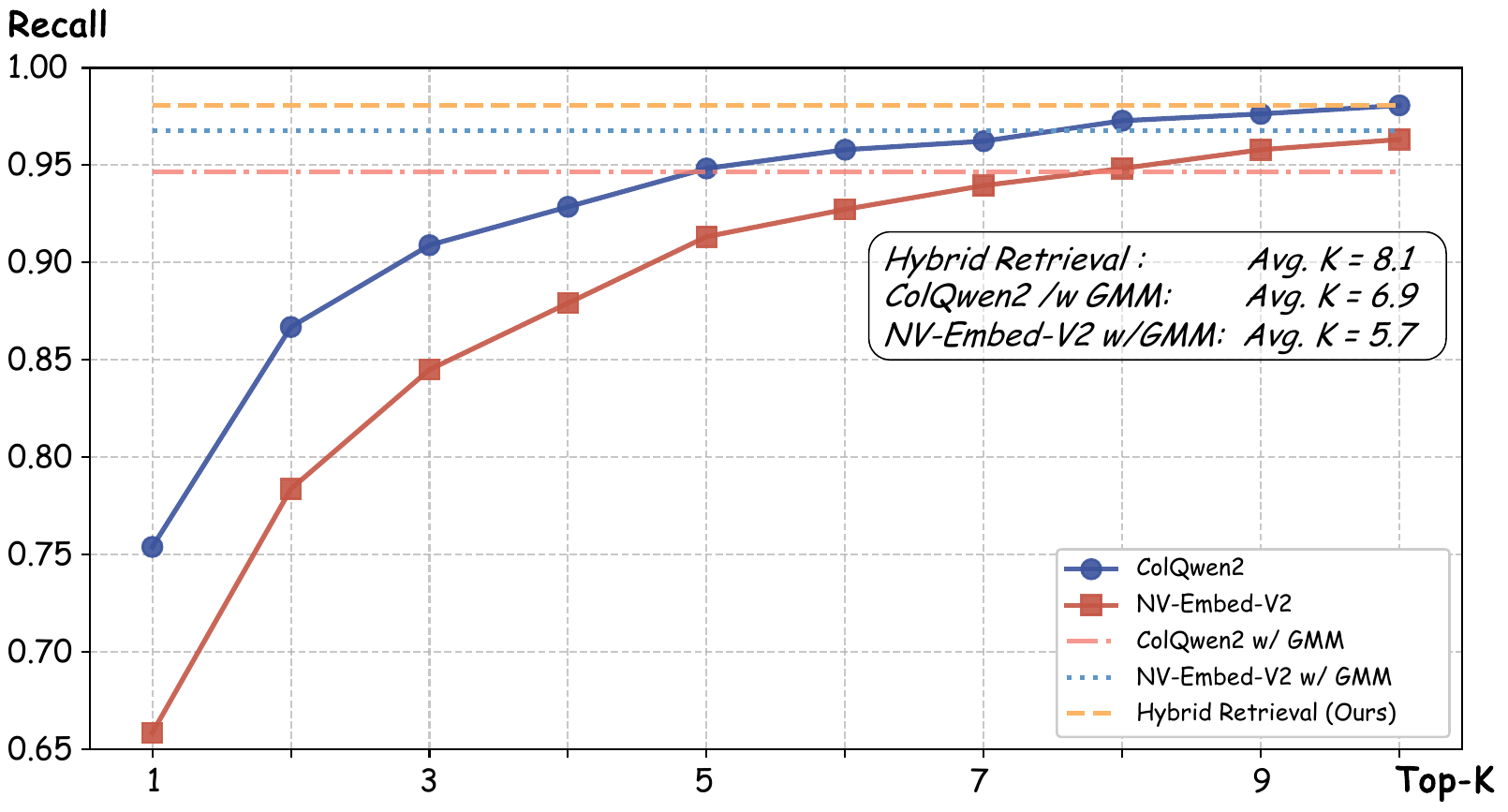}
    \caption{Retrieval performance across different retrievers and hybrid retrieval, along with ablations on GMM. 
    }
    \label{fig:ret}
\end{figure}

\subsection{Retrieval Evaluation}
In Table \ref{tab:ret}, we report the detailed performance for various retrievers, including OCR-based and visual-based. 
Due to the uncertainty of dynamical retrieval across queries, we use the average length of results for analysis.
Our goal is to incorporate more relevant information within a shorter context while minimizing the impact of noise and reducing computational cost without losing valuable information. 
Dynamic retrieval can achieve better recall performance with a smaller context length, while hybrid retrieval combines the results of two pipelines achieving state-of-the-art performance.

\section{Analysis}
\subsection{Ablations}
Table \ref{tab:ablation} presents the impact of different retrievers and generation methods on performance. We have decomposed the dynamic retrieval into two components, Dynamic and Hybrid. Naive refers to the method of direct input, which is most commonly used as baselines. Dynamic indicates using GMM to fit the optimal recall distribution based solely on the visual pipeline. Hybrid refers to merging the visual and the textual retrieval results directly, which leads to suboptimal results due to long contexts. Experiments demonstrate that the effectiveness and scalability of our improvements on retrieval and generation modules, as well as their combination, can comprehensively enhance end-to-end performance from various perspectives.

\begin{table}[!t]
    \small
    \centering
    \caption{Ablation study on ViDoSeek benchmark.}
    \label{tab:ablation}
    \resizebox{0.85\linewidth}{!}{
    \begin{tabular}{ccc|cc|c}
    \toprule
    \multicolumn{3}{c|}{\textsc{\textbf{Retrieval}}} & \multicolumn{2}{c|}{\textsc{\textbf{Generation}}} & \multirow{2}{*}{\textbf{Accuracy}} \\
    \textbf{Naive} & \textbf{Dynamic} & \textbf{Hybrid} & \textbf{Naive} &\textbf{Multi-Agent} &  \\
    \midrule
     \ding{51} & & & \ding{51} & & 72.1 \\
     & \ding{51} & & \ding{51} & & 72.8 \\
     & & \ding{51} & \ding{51} & & 74.1 \\
     & \ding{51} & \ding{51} & \ding{51} & & 74.3\\
     \ding{51} & & & & \ding{51} & 77.3 \\
     & \ding{51} & \ding{51} & & \ding{51} & \textbf{79.4}\\
    \bottomrule
    \end{tabular}
}
\end{table}

\subsection{Time Efficiency}
\label{sec:analysis:time}
\paragraph{How does dynamic retrieval balance latency and accuracy?}
In traditional RAG systems, using a small top-K value may result in missing critical information, whereas employing a larger value can introduce noise and increase computational overhead. ViDoRAG dynamically determines the number of documents to retrieve based on the similarity distribution between the query and the corpus. This approach ensures that only the most relevant documents are retrieved, thereby reducing unnecessary computations from overly long contexts and accelerating the generation process. As shown in Table \ref{tab:time_recall}, we compare retrieval with and without GMM based on the Naive method. The experiments indicate that GMM may reduce recall due to distribution bias. However, because it significantly shortens the generation context, it effectively improves performance in end-to-end evaluations.
\begin{table}[!t]
    \small
    \centering
    \caption{Evaluation of Dynamic Retrieval Methods.}
    \resizebox{0.71\linewidth}{!}{
    \label{tab:time_recall}
    \begin{tabular}{lcc}
    \toprule
    \textbf{Method} & \textbf{Accuracy} $\uparrow$ & \textbf{Avg. Pages} $\downarrow$\\
    \midrule
    w/o GMM &  72.1 & 10\\
    w/ GMM & \textbf{72.8} & \textbf{6.76} \\
    \bottomrule
    \end{tabular}
}
\end{table}

\begin{figure}[!h]
    \centering 
    \includegraphics[width=0.93\linewidth]{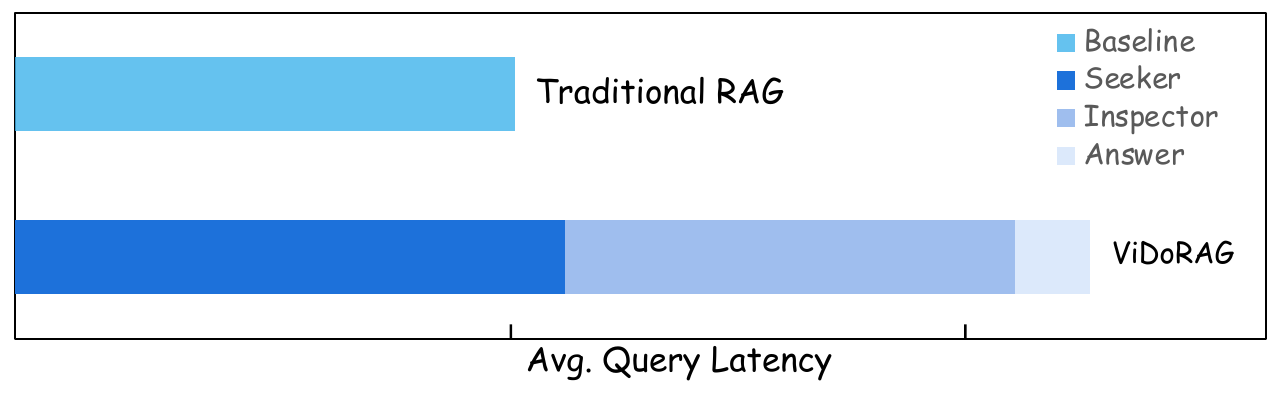}
    \caption{Latency Analysis on Generation.}
    \label{fig:gen_time}
\end{figure}

\paragraph{Latency Analysis of the Multi-Agent Generation.}
There is an increase in delay due to the iterative nature of the multi-agent system, as shown in Fig. \ref{fig:gen_time}. Each agent performs specific tasks in a sequential manner, which adds a small overhead compared to traditional straightforward RAG. However, despite the increase in latency, the overall performance improves due to the higher quality of generated answers, making the trade-off between latency and accuracy highly beneficial for complex RAG tasks.

\subsection{Modalities and Strategies of Generation}
\begin{figure}[!h]
    \centering 
    \includegraphics[width=0.99\linewidth]{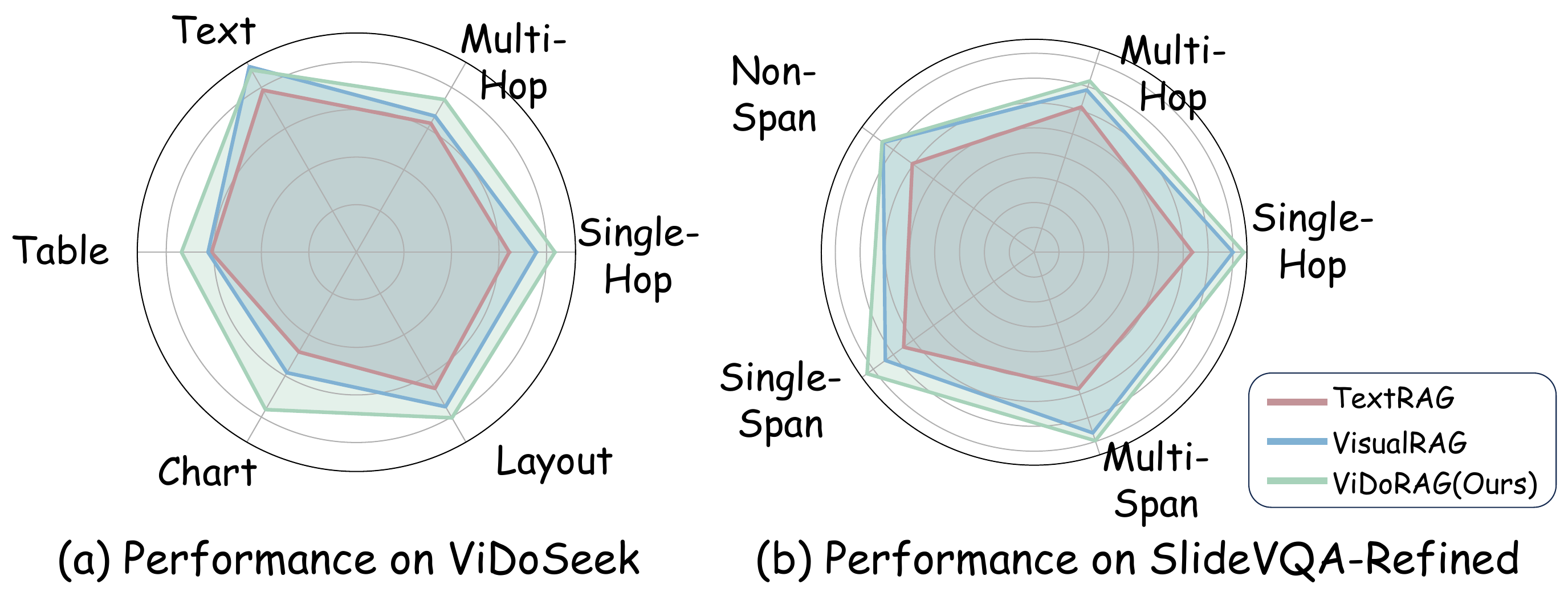}
    \caption{Performance across different types of queries on our ViDoSeek and the refined SlideVQA datasets.}
    \label{fig:compare_radar}
\end{figure}

As shown in Fig. \ref{fig:compare_radar}, the vision-based pipeline outperforms the text-based pipeline across all types, even for queries related to text content. Generally speaking, due to models' inherent characteristics, the reasoning ability of LLMs is stronger than that of VLMs. However, the lack of visual information makes it difficult for models to identify the intrinsic connections between pieces of information. This also poses a challenge for the generation of content based on visually rich documents. While obtaining visual information, VidoRAG further enhances the reasoning capabilities of VLMs, striking a balance between accuracy and computational load.

\begin{figure}[!h]
    \centering 
    \includegraphics[width=0.91\linewidth]{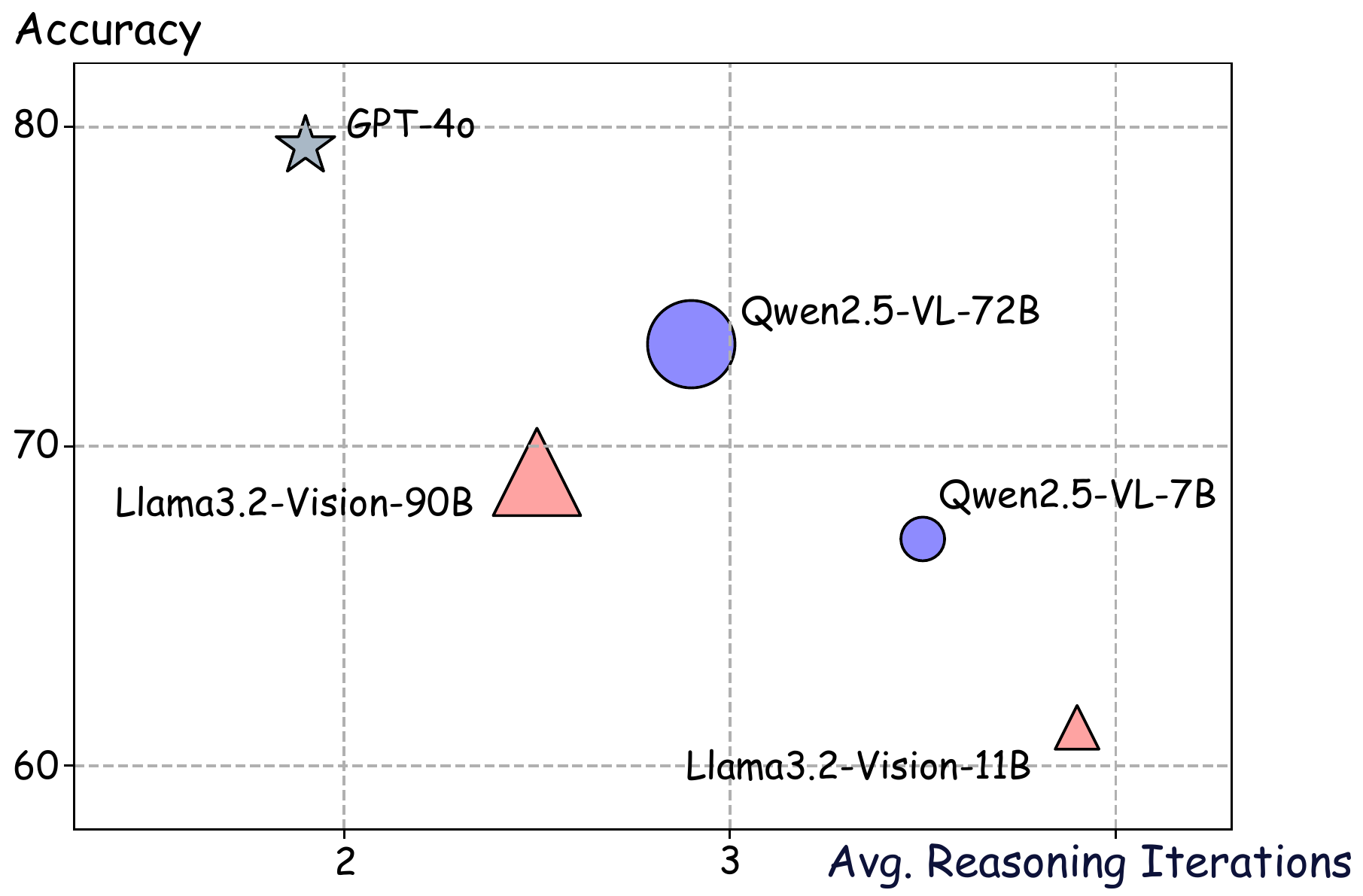}
    \caption{Scaling behavior with ViDoRAG.}
    \label{fig:scale}
\end{figure}

\subsection{Performance with Test-time Scaling}

Fig. \ref{fig:scale} illustrates the number of interaction rounds between the seeker and inspector within ViDoRAG based on different models. 
Due to the limited instruction capabilities of some models, we sampled 200 queries for the experiment.
Models with stronger performance require fewer reasoning iterations, while weaker models often need additional time to process and reach a conclusion.
Conditioning the model on a few demonstrations of the task at inference time has been proven to be a computationally efficient approach to enhance model performance\cite{brown2020language,min2021metaicl}. 
The results indicate that predefining tasks and breaking down complex tasks into simpler ones is an effective method for scaling inference.

\section{Conclusion}
In this work, we introduced ViDoRAG, a novel multi-agent RAG framework tailored for visually rich documents. By proposing a coarse-to-fine reasoning process and a multi-modal retrieval strategy, ViDoRAG significantly outperforms existing methods, achieving new SOTA on the ViDoSeek benchmark. Future work will focus on further optimizing the framework's efficiency while maintaining high accuracy, and exploring its potential in diverse real-world applications, such as education and finance, where visually rich document RAG is crucial.

% \clearpage
\section*{Limitations}
In addition to the advanced improvements mentioned above, our work has several limitations:  
\textbf{(1) Potential Bias in Query Construction.} The queries in ViDoSeek were constructed by human experts, which may introduce bias in the types of questions and the way they are phrased. This could affect the model's ability to handle more diverse and natural language queries from real-world users.
\textbf{(2) Computational Overhead of ViDoRAG.} The multi-agent framework, while effective in enhancing reasoning capabilities, introduces additional computational overhead due to the iterative interactions between the seeker, inspector, and answer agents. This may limit the scalability of the framework in scenarios with strict latency requirements.
\textbf{(3) Model Hallucinations.} Despite the improvements in retrieval and reasoning, the models used in ViDoRAG can still generate hallucinated answers that are not grounded in the retrieved information. This issue can lead to incorrect or misleading responses, especially when the model is overconfident in its generated content.

In summary, while ViDoRAG demonstrates significant improvements in visually rich document retrieval and reasoning, there are still areas for further enhancement, particularly in terms of generalization to diverse document types, reducing potential biases in query construction, optimizing the computational efficiency of the multi-agent framework, and addressing the issue of model hallucinations. Future work will focus on addressing these limitations to further improve the robustness and applicability of the model.

\section*{Ethical Considerations}

Our data does not contain any private or sensitive information, and all content is derived from publicly available sources. Additionally, the construction and refinement of the dataset were conducted in a manner that respects copyright and intellectual property rights.

% \section*{Acknowledgments}

% \newpage
\bibliography{custom}
\newpage
\clearpage
\appendix

\definecolor{lightgray}{gray}{0.95}
\definecolor{deepblue}{RGB}{70,130,180}
\definecolor{deepgray}{RGB}{119,136,153}
\lstdefinestyle{prompt}{
    basicstyle=\ttfamily\fontsize{7pt}{8pt}\selectfont,
    frame=none,
    breaklines=true,
    backgroundcolor=\color{lightgray},
    breakatwhitespace=true,
    breakindent=0pt,
    escapeinside={(*@}{@*)},
    numbers=none,
    numbersep=5pt,
    xleftmargin=5pt,
    aboveskip=2pt,
    belowskip=2pt,
}
\tcbset{
  aibox/.style={
    top=10pt,
    colback=white,
    % colframe=black,
    % colbacktitle=black,
    enhanced,
    center,
    % attach boxed title to top left={yshift=-0.1in,xshift=0.15in},
    % boxed title style={boxrule=0pt,colframe=white,},
  }
}
\newtcolorbox{AIbox}[2][]{aibox, title=#2,#1}

\section{Additional Experiments Details}

\paragraph{Backbones.} To thoroughly validate the effectiveness of ViDoRAG, we conducted experiments on various models across various baselines, including both closed-source and open-source models: GPT-4o, Qwen2.5-7B, Llama3.2-3B, Qwen2.5-VL-7B\cite{yang2024qwen2}, Llama3.2-Vision-90B. For OCR-based pipelines, we use PPOCR\cite{ma2019paddlepaddle} to recognize text within documents. Optionally, VLMs can also be employed for text recognition, as their OCR capabilities are quite strong.

\paragraph{Experimental Environments.}
We conducted our experiments on a server equipped with 8 A100 GPUs and 96 CPU cores. Open-source models require substantial computational resources.

\paragraph{Retrieval Implementation Details.} Due to the context length limitations of the model, we use the Top-$2K$ pages to fit the GMM and we restrict the output chunks of the GMM algorithm to be between $K/2$ and $K$, we set $K=10$ in practice. 

\section{More Details on Datasets}
\label{appendix:dataset_composition}

\subsection{Annotation Case}
\lstdefinestyle{mystyle}{
    language=Python,
    % commentstyle=\color{codegreen},
    % keywordstyle=\color{magenta},
    % numberstyle=\tiny\color{codegray},
    % stringstyle=\color{codepurple},
    basicstyle=\ttfamily \lst@ifdisplaystyle\tiny\fi,
    breakatwhitespace=false,
    breaklines=true,
    captionpos=b,
    keepspaces=true,
    numbers=left,
    numbersep=5pt,
    xleftmargin=0pt,  % 去除左侧缩进
    showspaces=false,  % 不显示空格
    showstringspaces=false,  % 不显示字符串中的空格
    showtabs=false,
    tabsize=2,
    columns=flexible,  % 使列宽自动调整
    moredelim=[is][\bfseries]{<highlight>}{</highlight>}
}

\begin{figure}[!h]
\begin{tcolorbox}[title={\textbf{\small Annotated Data Format}}]
{\small
\begin{lstlisting}[style=mystyle]
## JSON Format
{
    "uid": "04d8bb0db929110f204723c56e5386c1d8d21587_2",
    "query": "What is the temperature of Steam explosion of Pretreatment for Switchgrass and Sugarcane bagasse preparation?",
    "reference_answer": "195-205 Centigrade",
    "meta_info": {
        "file_name": "04d8bb0db929110f204723c586c1d8d21587.pdf",
        "reference_page": [
            10
        ], # may contain multiple pages
        "source_type": "2d_layout",
        "query_type": "Multi-Hop"
    }
}
\end{lstlisting}
}
\end{tcolorbox}
\caption{Annotation case in ViDoSeek.}
% \label{fig:annotated}
\end{figure}

\subsection{Details on ViDoSeek}
\paragraph{More Dataset Statistics.}
The statistical about ViDoSeek is presented in Table \ref{tab:data_statistic_slide}. We categorize queries from a logical reasoning perspective into single-hop and multi-hop. Text, Table, Chart and Layout represent different sources of reference.
\begin{table}[!h]
    \small
    \centering
    \caption{\textbf{Statistics of ViDoSeek.}}
    % \resizebox{1.0\textwidth}{!}{
    \label{tab:data_statistic}
    \begin{tabular}{lcc}
    \toprule
    \textsc{\textbf{Statistic}} & \textsc{\textbf{Number}}\\
    \midrule
    Total Questions & 1142\\
    \midrule
    Single-Hop & 645 \\
    Multi-Hop & 497 \\
    \midrule
    Pure Text & 80\\
    Chart & 157 \\
    Table & 175 \\
    Layout & 730\\
    \bottomrule
    \end{tabular}
% }
\end{table}
\paragraph{Dataset Difficulty.}
ViDoSeek sets itself apart with its heightened difficulty level, attributed to the multi-document context and the intricate nature of its content types, particularly the Layout category. The dataset contains both single-hop and multi-hop queries, presenting a diverse set of challenges. Consequently, ViDoSeek serves as a more comprehensive and demanding benchmark for RAG systems compared to previous works.

\subsection{Details on SlideVQA-Refined}
\paragraph{Dataset Statistics.}
We supplemented our experiments with the SlideVQA dataset to demonstrate the scalability of our method. 
SlideVQA categorizes queries from a logical reasoning perspective into single-hop and multi-hop. 
Non-span, single-span, and multi-span respectively refer to answers derived from a single information-dense sentence, reference information that is sparse but located on the same page, and reference information distributed across different pages.
The statistical information about dataset is presented in Table \ref{tab:data_statistic_slide}.

\begin{table}[!h]
    \small
    \centering
    \caption{\textbf{Statistics of SlideVQA-Refined.}}
    % \resizebox{1.0\textwidth}{!}{
    \label{tab:data_statistic_slide}
    \begin{tabular}{lcc}
    \toprule
    \textsc{\textbf{Statistic}} & \textsc{\textbf{Number}}\\
    \midrule
    Total Questions & 2020\\
    \midrule
    Single-Hop & 1486 \\
    Multi-Hop & 534 \\
    \midrule
    Non-Span & 358\\
    Single-Spin & 1347 \\
    Multi-Span & 315 \\
    \bottomrule
    \end{tabular}
% }
\end{table}
\paragraph{Dataset Difficulty.} The SlideVQA dataset focuses on evaluating the RAG system's ability to understand both visually sparse and visually dense information. When multi-hop questions involve reference information spread across different pages, it presents a significant challenge to the RAG system, further demonstrating the effectiveness of our approach.

\section{Data Construction Details}
\label{appendix:data_construction_pipeline}

To construct the ViDoSeek dataset, we developed a four-step pipeline to ensure that the queries meet our requirements. 
\paragraph{Step 1. Document Collecting.}
We collected English-language slides containing 25 to 50 pages, covering 12 domains such as economics, technology, literature, and geography, etc.

\paragraph{Step 2. Query Creation.}

To make the queries more suitable for RAG over a large-scale collection, our experts constructed queries based on the following requirements: (\romannumeral1) Each query must have a unique answer when paired with the document. (\romannumeral2) The query must include unique keywords that point to the specific document and pages. (\romannumeral3) The query should require external knowledge. Additionally, we encouraged constructing queries in various forms and with different sources and reasoning types to better reflect real-world scenarios. Our queries not only focus on types of references, including text, tables, charts, and layouts, but also provide a classification of reasoning types, including single-hop and multi-hop.

\paragraph{Step 3. Quality Review.}
To effectively evaluate the generation and retrieval quality of our RAG system, we require queries that yield unique answers, preferably located on a specific page or within a few pages. However, in large-scale retrieval and generation tasks, relying solely on manual annotation is challenging due to human cognitive limitations. To address this, we propose a review module that automatically identifies problematic queries. This module consists of two steps: (\romannumeral1) We prompt LLMs to filter out queries that may have multiple answers across the document collection; for example, the question \emph{What is the profit for this company in 2024?} might have a unique answer within a single document but could yield multiple answers in a multi-document setting. (\romannumeral2) For the remaining queries, we retrieve the top-\emph{k} slides for each query and use a VLM to determine whether each slide can answer the query. If only the golden page can answer the question, we consider it to meet the requirements. If pages other than the golden page can answer the query, we have experts manually evaluate and refine them.

\paragraph{Step 4. Multimodal Refine.}
% refine问题 同时保证问题中不含答案
In this final step, we refine the queries that did not meet our standards during the quality review. The goal is to adjust these queries so they satisfy the following requirements: (\romannumeral1) The refined query should point to specific pages within the large collection with minimal additional information; (\romannumeral2) The refined query must retain its original meaning. 
We use carefully designed VLM-based agents to assist us throughout the entire dataset construction pipeline. The prompt is presented in Fig. \ref{fig: reviewer} and Fig. \ref{fig: multi_reviewer}, respectively. We will first perform filtering based on semantics, and then conduct a fine-grained review using a multimodal reviewer.

\section{More Details about Multi-Agent Generation with Iterative Reasoning}
\label{appendix: gen}
We designed prompts to drive VLMs-based agents, and through our experiments, we found that some open-source models require the design of few-shot examples to learn specific thought patterns. See detailed prompts in Fig. \ref{fig: seeker}, Fig.\ref{fig: inspector} and Fig.\ref{fig: answer}.

\begin{figure*}[!ht] 
\begin{AIbox}{Query Reviewer Prompt.}
{\color{black}\bf \large System Prompt:} 
\vspace{1mm}
\\
\textbf{Task}  \\
I have some QA data here, and you can observe that the questions can be divided into two categories:\\
The category \#A: When you see this question alone without a given document, you are sure to find a unique document in a corpus to provide a unique answer. The question having some key words to help you locate the document from corpus.\\
The category \#B: When you see this question alone without a given document, you will find hard to locate a document to give a deterministic answer for this question, because you will find multiple candidate documents in a corpus, which may lead to different answers for this question. The question do not have any special key words to help you locate the document from corpus.

\textbf{Examples}\\
The number mentioned on the right of the leftside margin? \#B\\
What is the date mentioned in the second table? \#B\\
What is the full form of PUF? \#A\\
What is the number at the bottom of the page, in bold? \#B\\
Who presented the results on cabin air quality study in commercial aircraft? \#A\\
What is the name of the corporation? \#B\\
Which part of Virginia is this letter sent from? \#B\\
who were bothered by cigarette odors? \#A\\
which cigarette would be better if offered on a thicker cigarette? \#A\\
Cigarettes will be produced and submitted to O/C Panel for what purpose? \#A\\
What is the heading of first table? \#B\\
What is RIP-6 value for KOOL KS? \#A\\
Which test is used to evaluate ART menthol levels that has been shipped? \#A\\
How much percent had not noticed any difference in the odor of VSSS? \#A\\
What is the cigarette code of RIP-6(W/O Filter) 21/4SE? \#A\\
what mm Marlboro Menthol were subjectively smoked by the Richmond Panel? \#A\\
What are the steps of Weft Preparation between Spinning bobbin and Weaving? \#A\\
What level comes between Middle Managers and Non-managerial Employees? \#A\\
What are the six parts of COLLABORATION MODEL of the organization where James has a role of leading the UK digital strategy? \#A

\tcblower
{\color{black}\bf \large User Prompt:}\\
Query: {\color{deepblue}\bf \{Query Description\}} 
\end{AIbox}
\vspace{-1em}
\caption{Prompt of Query Reviewer.}
\label{fig: reviewer}
\end{figure*}

\begin{figure*}[!ht] 
\begin{AIbox}{Multi-Modal Reviewer Prompt.}
{\color{black}\bf \large System Prompt:} 
\vspace{1mm}
\\
Please check the image, tell me whether the image can answer my question.

\tcblower
{\color{black}\bf \large User Prompt:}\\
Query: {\color{deepblue}\bf \{Query Description\}}\\
Image: {\color{deepblue}\bf \{Relevant Image\}} 
\end{AIbox}
\vspace{-1em}
\caption{Prompt of Multi-Modal Reviewer.}
\label{fig: multi_reviewer}
\end{figure*}

\begin{figure*}[!ht] 
\begin{AIbox}{Multi-Modal Query Refiner Prompt.}
{\color{black}\bf \large System Prompt:} 
\vspace{1mm}
\\
\textbf{Task}  \\
Rewrite the following question so that it contains specific keywords that clearly point to the provided document, ensuring that it would likely match this document alone within a larger corpus.\\
\\
\textbf{Instruction}\\
- Do not add any additional information or context to the question.\\
- You should not change the meaning of the question.\\
- If the question is already specific and unique, you may leave it unchanged.\\
- Please make the sentences you have rewritten more diverse and fluent. \\
\\
\textbf{Examples}\\
- Original question: GIS data integration is part of which process?\\
- Rewritten question: Citizen Science shows which process the GIS data integration is part of?\\

- Original question: What percentage of apps ranked in the top five for including what resulted in a 10,3\% Ranking Increase?\\
- Rewritten question: According to the App Store Optimization what percentage of apps ranked in the top five for including what resulted in a 10,3\% Ranking Increase?\\

- Original question: Who is the author of the book, the title of which is the same as the section title of the presentation?\\
- Rewritten question: Who is the author of the book, the title of which is the same as the section title of the presentation by Michael Sahota and Olaf Lewitz?\\

- Original question: Which region of the world accounts for the highest percentage of revenues in the year 12\% GROWTH is achieved?\\
- Rewritten question: Which region of the world accounts for the highest percentage of revenues in the year 12\% GROWTH is achieved?\\

- Original question: What directly follows "conduct market research to refine" in the figure?\\
- Rewritten question: What directly follows "conduct market research to refine" in the figure within the Social Velocity Strategic Plan Process?\\

- Original question: How can the company which details 24 countries in the report be contacted?\\
- Rewritten question: How can the company which details 24 countries in the Global Digital Statistics 2014 report, be contacted?\\

- Original question: What substances are involved in the feeding of substrates?\\
- Rewritten question: What substances are involved in the feeding of substrates during the production of penicillin?\\

\tcblower
{\color{black}\bf \large User Prompt:}\\
Query: {\color{deepblue}\bf \{Query Description\}} \\
Document: {\color{deepblue}\bf \{Document Description\}} \\
Image: {\color{deepblue}\bf \{Image File\}} \\

\end{AIbox}
\vspace{-1em}
\caption{Prompt of Multi-Modal Refiner.}
\label{fig: refiner}
\end{figure*}

\begin{figure*}[!ht] 
\begin{AIbox}{Seeker Agent Prompt.}
{\color{black}\bf \large System Prompt:} 
\vspace{1mm}
\\
\textbf{Character Introduction}  \\
You are an artificial intelligence assistant with strong ability to find references to problems through images. The images are numbered in order, starting from zero and numbered as 0, 1, 2 ... Now please tell me what information you can get from all the images first, then help me choose the number of the best picture that can answer the question. 

\textbf{Response Format}  \\
The number of the image is starting from zero, and counting from left to right and top to bottom, and you should response with the image number in the following format:
\begin{lstlisting}[style=prompt]
{
    "reason": Evaluate the relevance of the image to the question step by step,
    "summary": Extract the information related to the problem,
    "choice": List[int]
}
\end{lstlisting}

\textbf{Response Example}  \# open-source models sometimes need few-shot instructions.
\begin{lstlisting}[style=prompt]
Example 1: Question: Who is the person playing a musical instrument in restaurant? 
Response to Example 1: 
{
    "reason": "Image 0 shows that KFC on Renmin Road has a birthday party on February 3rd. I can know that there are musical instruments playing in Shanghai hotels during meals from Image 1. Image 2 shows that this is an invitation letter for the music performance of the New Year's Concert at Qintai Art Museum on December 31st. The question is related to the restaurant, and Image 2 is not relevant to the question.",
    "summary": "KFC on Renmin Road has a birthday party on February 3rd;Shanghai hotels have musical instruments playing during meals;The Qintai Art Museum will hold a New Year's concert on December 31st.",
    "choice": [0, 1]
}

Example 2: Question: What time is the train departing from hangzhou to beijing?
Response to Example 2:
{
    "reason": "Image 0 shows that Beijing has a temperature of 18 degrees Celsius. Image 0 is a train ticket from hangzhou to beijing showing a departure time of 14:30. Image 1 is a photo of a train station clock, but it's blurry and hard to read the exact time. Image 2 shows a train schedule with multiple departure times listed. Image 3 is the timetable of Hangzhou Xiaoshan International Airport, and this image is not related to the issue. I think Image 0 is the most relevant to the question.",
    "summary": "The train ticket shows a departure time of 14:30;The train station clock is blurry;Train schedule shows time.",
    "choice": [0]
}

Example 3: Question: Where can I find a bookstore that sells rare books? 
Response to Example 3: 
{
    "reason": "Image 0 is a street view of a shopping mall with various stores, but no bookstores are visible. Image 1 shows a sign for a bookstore called "Rare Finds Bookstore" specializing in rare books. Image 2 is a map with multiple bookstores marked, but it doesn't specify if they sell rare books. Image 3 is a photo of a library, which is not a place to buy books. Image 5 is a rare books list, which includes the names and prices of various books. ",
    "summary": "The shopping mall has no visible bookstores;Rare Finds Bookstore specializes in rare books;Map shows multiple bookstores but doesn't specify rarity;Library is not for buying books;The price list includes the prices and names of rare books.",
    "choice": [1, 5]
}
\end{lstlisting}

\tcblower
{\color{black}\bf \large User Prompt:}\\
Query: {\color{deepblue}\bf \{Query Description\}} \\
Images: {\color{deepblue}\bf \{Candidate Images\}} \\
Reflection:  {\color{deepblue}\bf \{Feedback From Inspector\}} 

\end{AIbox}
\vspace{-1em}
\caption{Prompt of Seeker Agent.}
\label{fig: seeker}
\end{figure*}

\begin{figure*}[!ht] 
\begin{AIbox}{Inspector Agent Prompt.}
{\color{black}\bf \large System Prompt:} 
\vspace{1mm}
\\
\textbf{Character Introduction}  \\
You are an artificial intelligence assistant with strong ability to answer questions through images. Please provide the answer to the question based on the information provided.

\textbf{Task Description}  \\
- If the images can answer the question, please answer the question directly.\\
- If the images are not enough to answer the question, please tell me which pictures are related to the question.

\textbf{Response Format}  \\
- If the images can answer the question, please answer the question directly:
\begin{lstlisting}[style=prompt]
{
    "reason": Solve the question step by step,
    "answer": Answer the question briefly with several words,
    "reference": List[int]
}
\end{lstlisting}

- If the images are not enough to answer the question, please tell me what additional information you need, and tell me which pictures are related to the question:
\begin{lstlisting}[style=prompt]
{
    "reason": Evaluate the relevance of the image to the question one by one, and solve the question step by step,
    "information": Carefully clarify the information required,
    "choice": List[int]
}

\end{lstlisting}

\textbf{Response Example}  \# open-source models sometimes need few-shot instructions.
\begin{lstlisting}[style=prompt]
- Example 1:
{
    "reason": "The image only provides information about the Bohr Model and does not include details about subshells in the Modern Quantum Cloud Model.",
    "information": "More information about the Bohr Model.",
    "choice": []
}

- Example 2:
{
    "reason": "The images provide information about the #swallowaware campaign, including its aims and how they were measured. However, specific details on the success metrics are not clearly visible in the provided images.",
    "information": "More information about the success metrics of the #swallowaware campaign.",
    "choice": [0, 1]
}

- Example 3:
{
    "reason": "We first found the restaurant name on the menu, and then we located the restaurant in the city center on the map.",
    "answer": "city center",
    "reference": [2, 3]
}

- Example 4:
{
    "reason": "The entire process, from input, processing to output, ultimately produces a product with a purity of 42%.",
    "answer": "42%",
    "reference": [0]
}
\end{lstlisting}

\tcblower
{\color{black}\bf \large User Prompt:}\\
Query: {\color{deepblue}\bf \{Query Description\}} \\
Plan: {\color{deepblue}\bf \{Thought From Last Step.\}} \\
Images: {\color{deepblue}\bf \{Images Pending Review.\}} 

\end{AIbox}
\vspace{-1em}
\caption{Prompt of Inspector Agent.}
\label{fig: inspector}
\end{figure*}

\begin{figure*}[!ht] 
\begin{AIbox}{Answer Agent Prompt.}
{\color{black}\bf \large System Prompt:} 
\vspace{1mm}
\\
\textbf{Character Introduction}  \\
You are an artificial intelligence assistant with strong ability to answer questions through images. Please provide the answer to the question based on the information provided and tell me which pictures are your references.

\textbf{Response Format}  \\
Please provide the answer in JSON format:
\begin{lstlisting}[style=prompt]
{
    "reason": Solve the question step by step,
    "answer": Answer the question briefly with several words,
    "reference": List[int]
}
\end{lstlisting}

\tcblower
{\color{black}\bf \large User Prompt:}\\
Query: {\color{deepblue}\bf \{Query Description\}} \\
Draft Answer: {\color{deepblue}\bf \{Draft Answer From Inspector\}} \\
Images: {\color{deepblue}\bf \{Reference Images\}}

\end{AIbox}
\vspace{-1em}
\caption{Prompt of Answer Agent.}
\label{fig: answer}
\end{figure*}

\end{document}